\documentclass[conference]{IEEEtran}
\usepackage{cite}
\usepackage{amsmath,amssymb,amsfonts}
\usepackage{algorithmic}
\usepackage{graphicx}
\usepackage{textcomp}
\usepackage{xcolor}
\usepackage{hyperref}
\usepackage{tikz}
\usepackage{graphicx}
\usepackage{booktabs}
\usepackage{bm}
\usepackage{svg}
\usepackage{adjustbox}
\usepackage{svg-extract}

\usetikzlibrary{shapes.geometric, arrows.meta, positioning, calc, fit, backgrounds}

\tikzset{
    box/.style={rectangle, rounded corners, minimum width=3cm, minimum height=1cm, text centered, draw=black, font=\small},
    sim/.style={rectangle, rounded corners, minimum width=3.5cm, minimum height=1.2cm, text centered, draw=black, fill=orange!15, font=\small},
    metrics/.style={rectangle, rounded corners, minimum width=3.5cm, minimum height=1.2cm, text centered, draw=black, fill=purple!10, font=\bfseries\small},
    agent_container/.style={draw=black, fill=green!5, rounded corners, inner sep=15pt, thick, dashed},
    arrow/.style={-Stealth, thick}
}

\title{Breaking the Filter Bubble: A Semantic Pareto-DQN Framework for Multi-Objective Recommendation}

\author{%
  \IEEEauthorblockN{Cláudio Lúcio Do Val Lopes}
  \IEEEauthorblockA{A3Data - MG, Brazil \\
    claudio.lucio@a3data.com.br}
  \and
  \IEEEauthorblockN{Lucca Machado da Silva}
  \IEEEauthorblockA{A3Data - MG, Brazil\\
    lucca.machado@a3data.com.br}
  \and
  \IEEEauthorblockN{André de Oliveira Brandão}
  \IEEEauthorblockA{A3Data - MG, Brazil \\
    andre.brandao@a3data.com.br}
}

\begin{document}

\maketitle

\begin{abstract}
Recommender systems often induce filter bubbles and semantic homogenization by monolithically optimizing for immediate user engagement. Standard single-objective models, including traditional Deep Q-Networks, are ill-equipped to navigate the trade-offs between platform retention and critical societal values like information diversity and provider fairness. To address these limitations, we introduce a multi-objective reinforcement learning framework that formalizes recommendation as a semantic multi-objective Markov decision process. By integrating high-fidelity semantic embeddings with a Pareto-DQN agent, our architecture treats engagement, diversity, and fairness as distinct, non-aggregable reward signals, avoiding the pitfalls of static reward scalarization. Empirical evaluations on the MovieLens small dataset shows that our hypervolume based action selection disrupts the feedback loops responsible for semantic collapse. By sustaining high state-trajectory variance, the Pareto-DQN effectively maps the Pareto frontier, achieving  gains in auxiliary societal objectives with only marginal impacts on engagement. This work provides a path toward intrinsically aligned, responsible recommender systems.
\end{abstract}

\begin{IEEEkeywords}
AI Alignment, Many-Objective Optimization, Recommender Systems, Responsible AI, Reinforcement Learning, Filter Bubbles.
\end{IEEEkeywords}

\section{Introduction}

Recommender systems serve as the primary gatekeepers of digital information, aiming to generate personalized suggestions tailored to individual preferences \cite{Roy2022}. However, these systems have been empirically shown to progressively narrow the content users are exposed to, creating filter bubbles that severely reduce diversity over time \cite{Nguyen2014FilterBubble}. Standard single-objective models, including traditional Deep Q-Networks (DQN) \cite{QADER2025}, typically optimize strictly for immediate user engagement.  This induces semantic homogenization, making single-objective architectures ineffective at balancing engagement with broader, structurally antagonistic societal values such as information diversity and provider fairness \cite{Ge2022Pareto}.

To address the structural limitations of single-objective optimization, we present a Multi-Objective Reinforcement Learning (MORL) framework that combines a semantic embedding approach with Pareto Deep Q-Learning (Pareto-DQN) to successfully navigate these inherent trade-offs \cite{Hayes2022}. Our pipeline constructs a semantic state representation leveraging the \texttt{all-MiniLM-L6-v2} Sentence Transformer \cite{Reimers2019SBERT}. Co-locating users and items within the same high-fidelity latent space ensures that standard inner products natively capture semantic relevance \cite{Zhang2019}, while also enabling zero-shot generalization to unseen items, thereby mitigating the collaborative filtering cold-start problem. 

Through empirical evaluations on the MovieLens dataset \cite{Harper2015}, we quantify filter bubble mitigation via user embedding variance and analyze the \textit{price of responsibility}. Our results show that the Pareto-DQN effectively disrupts the positive feedback loops that drive semantic collapse. By comprehensively mapping the Pareto frontier, the agent unlocks relative gains in diversity and fairness for only fractional, manageable drops in engagement, providing a highly scalable path toward intrinsically aligned, responsible recommender systems.

The remainder of this paper is organized as follows: Section \ref{sec:related_works} reviews the traditional single-objective and multi-objective in recommender systems reinforcement learning. Section \ref{sec:methodology} formalizes the recommendation task as a semantic multi-objective Markov decision process, and details the architectural of our proposed agent. Section \ref{sec:experiments} presents the empirical evaluation, providing a analysis of convergence, the disruption of filter bubbles via semantic variance, and the empirical price of responsibility. Finally, Section \ref{sec:conclusion} concludes the study and outlines directions for future research.
\section{Related Works}\label{sec:related_works}

Reinforcement Learning (RL) has emerged as the state-of-the-art paradigm by formulating recommendations as a Markov Decision Process. This enables systems to optimize long-term engagement through continuous policy updates driven by real-time feedback \cite{Afsar2022}. Deep Reinforcement Learning (DRL) utilizing Deep Q-Networks (DQN) has become the value-based standard due to robust update strategies and good performance in discrete action spaces \cite{QADER2025}. Recent empirical evidence suggests that RL-based methods surpass supervised learning approaches because of their interactive nature and autonomous adaptation to dynamic user preferences \cite{Wang2022DRL}.

Despite these advances, traditional single-objective optimization represents a fundamental limitation. Real-world scenarios inherently involve multiple conflicting objectives requiring simultaneous optimization \cite{Rehman2025}. Standard DRL typically struggles to balance accuracy with non-accuracy metrics, making it ineffective for navigating complex real-world trade-offs \cite{Rehman2025}. Multi-Objective Reinforcement Learning (MORL) addresses this gap through the use of vectorial rather than scalar rewards, enabling to handle conflicting objectives \cite{Hayes2022}. This approach is grounded in Pareto optimization, a theoretical framework in which improving one objective typically cannot be achieved without altering another \cite{Hayes2022}. Recent industrial deployments demonstrate that Pareto-based DRL methods can simultaneously improve multiple business objectives while comprehensively modeling the complex relationships among them \cite{Ge2022Pareto}. Consequently, Pareto-DQN variants can optimize diversity, novelty, fairness, and engagement alongside traditional accuracy metrics \cite{Stamenkovic2022Diverse}.

Pareto-DRL represents the core of our proposal. Current applications of MORL span a diverse range of domains, including trip recommendation \cite{Chen2023MOTrip}, short video platforms \cite{Cai2023TwoStage}, electric vehicle charging \cite{Zhang2021EV}, educational systems \cite{Ren2023Education}, retention modeling \cite{Liu2024Sequential}, and fairness-utility trade-offs \cite{Ge2022Pareto}.

Our proposed framework integrates the advanced generative capabilities of Large Language Models (LLM), using a semantic space, into the recommendation pipeline. By employing a Pareto Deep Q-Network (Pareto-DQN) approach, the system effectively addresses the structural limitations of traditional collaborative filtering and the narrow focus of single-objective optimization. This methodology leverages the leading paradigm for responsible recommender systems by utilizing LLM embeddings as a high-fidelity mechanism to navigate complex value trade-offs, thereby bypassing the ethical risks and high costs associated with direct user experimentation.

\section{Methodology}\label{sec:methodology}

We present a MORL framework with a semantic approach applied to item recommendation that jointly optimizes user engagement, information diversity, and provider fairness. Recommender systems have been shown to progressively narrow the content users are exposed to, creating filter bubbles that reduce diversity over time \cite{Nguyen2014FilterBubble}. Our approach combines a semantic approach with Pareto deep Q-learning to navigate the inherent trade-offs between these conflicting objectives, mitigating this narrowing effect while maintaining recommendation quality.

\subsection{Semantic Embedding Pipeline}
To construct a robust state representation for our environment, we concatenate features as the title, genres, and user-generated tags into a single textual document $d_i$ for each item $ i$. This representation strategy has been empirically validated for content-based recommendation \cite{Zhang2019}. We map this document to a continuous vector space using the \texttt{all-MiniLM-L6-v2} Sentence Transformer \cite{Reimers2019SBERT}:

$$\bm{v_i} = \frac{\text{Encoder}(d_i)}{\|\text{Encoder}(d_i)\|_2}$$\label{eq:vi}

This model maps the text to a 384-dimensional vector optimized for similarity retrieval. We use $L_2$ normalization as a critical geometric constraint;
This inherently balances embedding quality with computational efficiency (the model contains 22M parameters and processes batches in $<50$ms on a standard CPU). By leveraging a pretrained semantic embedding space, we enable zero-shot generalization to unseen items, directly mitigating the cold-start problem common to purely collaborative filtering approaches. 

\subsection{Problem Formulation}

We formalize the sequential recommendation task as a Multi-Objective Markov Decision Process (MOMDP), defined by the tuple $\langle \mathcal{S}, \mathcal{A}, \mathcal{P}, \mathbf{R}, \gamma \rangle$. Here, $\mathcal{S}$ represents the state space of user preference profiles, $\mathcal{A}$ defines the action space of candidate items, $\mathcal{P}: \mathcal{S} \times \mathcal{A} \rightarrow \mathcal{S}$ dictates the deterministic transition dynamics of user preference drift, and $\gamma \in [0,1)$ is the discount factor. Crucially, the environment yields a vectorial reward function $\mathbf{R}: \mathcal{S} \times \mathcal{A} \times \mathcal{S} \rightarrow \mathbb{R}^3$, emitting a multi-dimensional signal capturing engagement, diversity, and fairness. Retaining this vectorial structure enables principled \textit{a posteriori} multi-objective optimization, circumventing the need for static, \textit{a priori} scalarization.

\subsubsection{State Space}

The state $\bm{s_t} \in \mathbb{R}^{384}$ encodes a user's preference profile as the centroid of their historically liked item embeddings:
$$s_t = \frac{1}{|H_u|} \sum_{i \in H_u} \bm{v_i}$$
where $H_u$ denotes the set of items rated $\geq 4.0$ by user $u$, and $\bm{v_i} \in \mathbb{R}^{384}$ is the $L_2$-normalized semantic embedding of item $i$. 

Unlike prior deep RL models, as in DRR-ave \cite{Liu2018DRR} that rely on auxiliary parameterized interaction modules, we directly deploy this centroid as a \textit{content-based user profile} \cite{Zhang2019}. Co-locating users and items within the same high-fidelity Sentence-BERT latent space \cite{Reimers2019SBERT} ensures that standard inner products natively yield semantic relevance. 

\subsubsection{Action Space}

Evaluating Q-values across the full catalog of over, $\mathcal{I} \ge 9,000$ items is computationally prohibitive. We employ a \textbf{stratified candidate pooling} to construct a bounded, dynamic action space $\mathcal{C}_t \subset \mathcal{A}$ of size $K=100$.

At each time step $t$, $\mathcal{C}_t$ is constructed by unifying two distinct sets: $K(1-\rho)$ items retrieved via cosine-distance nearest neighbor (KNN) search around the state centroid $s_t$, and $K\rho$ items injected from the exposure-sorted long tail, where $\rho = 0.3$. The combined candidate pool is subsequently shuffled to eliminate positional bias.

The injection parameter $\rho$ is theoretically critical for our MOO formulation. A purely KNN-based retrieval strategy would restrict the agent's action space entirely to high-affinity, already-popular items. Reserving a proportion $\rho$ of the action space for long-tail items, provides actionable pathways to optimize provider fairness. 

\subsubsection{Transition Dynamics}

To mathematically model the temporal evolution of user interests and the well-documented filter bubble effect \cite{Nguyen2014FilterBubble}, we implement a \textbf{preference drift} mechanism. Upon recommending item $a_t$ with embedding $v_{a_t}$, the user state transitions via a normalized exponential moving average:
$$s_{t+1} = \frac{(1 - \alpha) \bm{s_t} + \alpha \bm{v_{{a_t}}}}{\|(1 - \alpha) \bm{s_t} + \alpha \bm{v_{a_t}}\|_2}$$

Here, the hyperparameter $\alpha \in [0,1]$ governs the user's \textit{semantic susceptibility}. It dictates the step size of the state transition, controlling how a single interaction pulls the user's holistic preference centroid toward the recommended item's vector. This formulation embeds the filter bubble phenomenon directly into the MOMDP dynamics. If the policy greedily exploits high-affinity items to maximize immediate engagement, the recursive application of the $\alpha$-weighted update causes $s_t$ to rapidly collapse into a narrow semantic subspace. Consequently, the KNN-retrieved candidate pool $\mathcal{C}_{t+1}$ homogenizes, strictly degrading the agent's capacity to accrue future diversity rewards. 

\subsubsection{Vectorial Reward Function}

The environment emits a multi-objective reward vector $\bm{r_t} = [r_{eng}, r_{div}, r_{fair}]^\top \in \mathbb{R}^3$, where each component represents a distinct objective:

\textit{Engagement: } We map the semantic affinity  between the state and action embeddings through a sigmoid function:
$$r_{eng} = \sigma(\bm{s_t} \cdot \bm{v_{a_t}}) = \frac{1}{1 + e^{-(\bm{s_t} \cdot \bm{v_{a_t}})}}, $$
it bounds the reward to $(0, 1)$ and ensures smooth gradients for Q-value updates, assuming that semantic proximity drives user engagement.

\textit{Diversity: } measures semantic distance to prevent filter bubbles:
$$r_{div} = 1 - S_c(\bm{s_t} \cdot \bm{v_{a_t}}),$$
$S_c$ is the cosine similarity from the user's preference centroid
This directly incentivizes semantically distant recommendations, counteracting homogenization. 

\textit{Fairness: } To incentivize exploration of the long tail:
$$r_{fair} = \frac{1}{\log(1 + C(a_t))},$$
a monotonically decaying exposure reward, where $C(a_t) \ge 1$ tracks the cumulative platform exposure of item $a_t$. The denominator yields a steep exploratory gradient for novel items, providing a strong initial fairness signal without destabilizing the network's convergence.

\subsection{Standard DQN Agent (Baseline)}

We implement a single objective Deep Q-Network (Standard DQN). The reward value is just $r_{eng}$. During training, the agent ingests the concatenated state and candidate item embedding, $[\bm{s_t} \| \bm{v_{a_i}}] \in \mathbb{R}^{768}$, and is optimized via standard temporal difference learning solely on the $r_{eng}$ signal. Correspondingly, action selection evaluates candidates using a strictly greedy maximization of predicted engagement:
$$a_t^* = \arg\max_{a_i \in \mathcal{C}_t} Q(\bm{s_t}, a_i)$$

Because it employs a fixed, engagement-centric policy, the Standard DQN consistently drives the user state toward a narrow region of the embedding space. This structurally induces \textit{semantic homogenization}, the ``filter bubble'' effect, thereby providing a stark comparative baseline for evaluating the Pareto-DQN's capacity to explore the full trade-off surface and maintain recommendation diversity.

\subsection{Pareto-DQN Agent}

Our Pareto-DQN (PDQN) agent natively handles the multi-objective formulation through an item-centric architecture. Unlike traditional DQNs with fixed action outputs, our network ingests the concatenated user state and candidate item embedding, $[\bm{s_t} \| \bm{v_a}] \in \mathbb{R}^{768}$, enabling evaluation of arbitrary items via batched forward passes. The architecture bifurcates into two specialized Multi-Layer Perceptrons (MLPs):

\textbf{Reward Approximator ($\bar{R}$):} A 3-layer MLP (512-256-128 hidden units, ReLU activations) that predicts the immediate expected reward vector $\bar{\mathbf{r}} \in \mathbb{R}^3$. This network carries the primary learned signal and is optimized via Mean Squared Error (MSE) against the observed empirical reward vectors.

\textbf{Continuous Pareto Surface Estimation ($ND_t$):}  $ND_t$  functions as a continuous manifold estimator, treating the Pareto front as a conditional scalar function. Consider $d$, the dimensional reward vector ($\mathbf{r}_t = [r_{eng}, r_{div}, r_{fair}]^\top$). To mathematically decouple the immediate reward from the long-term expected return, we define $\mathbf{o} \in \mathbb{R}^3$ as the coordinates of the expected return space. The network takes the state $\bm{s_t}$, the action embedding $\bm{v_a}$, and $d-1$ additional values $\mathbf{o}_{1:2}$ corresponding to all but the last objective as input \cite{VanMoffaert2014PQL}. 

Specifically, the network ingests $[\bm{s_t} \parallel \mathbf{o}_{1:2} \parallel \bm{v_{a}}] \in \mathbb{R}^{770}$, where $\mathbf{o}_{1:2} = [o_{eng}, o_{div}]^\top \in [0,1]^2$ are uniformly sampled objective coordinates representing target returns for engagement and diversity. The output is the predicted maximum achievable return for the remaining $d$-th objective, fairness ($\hat{o}_{fair}$). Combining this prediction with the input coordinates yields a single point on the estimated Pareto surface. 

By drawing $n$ uniform samples $\{\mathbf{o}_{1:2}^{(1)}, \dots, \mathbf{o}_{1:2}^{(n)}\}$, we discretely reconstruct the continuous Pareto surface. The network parameters are optimized via Mean Squared Error (MSE).

\subsubsection{Action Selection via Hypervolume}

For each candidate action $a_i \in \mathcal{C}_t$, the set of expected returns is constructed by applying a vector-sum operation ($\oplus$) that adds the estimated immediate reward to each element of the non-dominated future returns:
$$Q_{set}(s_t, a_i) = \bar{\mathbf{r}}(s_t, v_{a_i}) \oplus \gamma ND_t(s_t, v_{a_i})$$
where $\gamma $ is the discount factor. 

To evaluate the quality of a given $Q_{set}$ and apply an $\epsilon$-greedy mechanism, we utilize a Hypervolume indicator. The hypervolume computes the total $d$-dimensional volume bounded by the points in $Q_{set}$ relative to a strict lower-bound reference point $\mathbf{r}_{ref}$, \cite{VanMoffaert2013HV, Li2024DeepPareto}. The greedy policy selects the action that maximizes this volume:
$$a_t^* = \arg\max_{a_i \in \mathcal{C}_t} \text{HV}(Q_{set}(s_t, a_i), \mathbf{r}_{ref})$$

Hypervolume is strictly Pareto-compliant; maximizing it mathematically guarantees convergence toward the true Pareto front without necessitating \textit{a priori} preference weighting. 

\section{Experiments}\label{sec:experiments}

To empirically validate our framework, we design an evaluation protocol centered on three core dimensions: (i) multi-objective trade-offs between engagement, diversity, and fairness; (ii) filter bubble mitigation, quantified via user embedding variance; and (iii) the \textit{Price of Responsibility}, defined as the engagement cost incurred by enforcing responsible optimization constraints. 

We utilize the MovieLens-Small dataset \cite{Harper2015}, comprising 100,836 ratings across 9,742 movies and 610 users, where filter bubble effects were empirically documented \cite{Nguyen2014FilterBubble}. To evaluate zero-shot generalization to unseen preference geometries, we partition the 609 qualifying users (those with ratings $\geq 4.0$) into 548 training users and 61 held-out test users using a strict 90/10 split. Test users are entirely excluded from the training environment.

We evaluate two specific agent architectures:
\begin{itemize}
    \item \textbf{Standard DQN (Baseline):} A single-objective agent optimizing strictly for engagement ($r_{eng}$). It is implemented as a 2-layer item-centric MLP (512-256 hidden units, ReLU activations) trained via standard MSE loss on scalar Q-values.
    \item \textbf{Pareto-DQN (Proposed):} Our multi-objective agent that optimizes the full vectorial reward $\mathbf{r}_t = [r_{eng}, r_{div}, r_{fair}]^\top$ via hypervolume-based action selection over continuous Pareto surface approximations.
\end{itemize}

Both agents perform 15,000 environment interactions (100 episodes of 150 steps). Training utilizes an item-centric replay buffer (capacity 5,000) and the Adam optimizer ($\text{lr} = 10^{-4}$), drawing uniform mini-batches of size 32. Target networks are updated via hard copies every 100 steps to stabilize temporal difference  learning. We employ an $\epsilon$-greedy exploration strategy ($\epsilon_0 = 1.0$, minimum $0.01$), decaying $\epsilon$ exponentially by 0.999 at the conclusion of each episode. 

Evaluation is conducted deterministically ($\epsilon = 0$) on the 61 held-out test users, running one full episode per user. To ensure statistical robustness and reproducibility, the entire pipeline is executed across five independent trials, with all reported metrics representing mean values $\pm$ standard deviations. The framework is implemented using PyTorch 2.0 and Gymnasium 0.29, relying on \texttt{pymoo} for exact hypervolume computations  \cite{pymoo}.

\subsection{Training Dynamics}

\begin{figure}[htpb]
\centering
\adjustbox{trim=0.22cm 0cm 0cm 0cm, clip}{
\includesvg[width=0.50\textwidth]{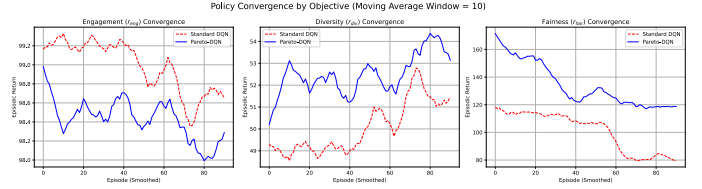}
}
\caption{Training convergence by objective (smoothed with a moving average window of 10). The Standard DQN (red dashed) aggressively maximizes engagement, leading to progressive fairness degradation and lower overall diversity. In contrast, the Pareto-DQN (blue solid) successfully navigates the trade-off surface, sustaining significantly higher fairness and diversity while incurring a marginal, controlled penalty in engagement.
\textit{Note: The y-axis utilizes a narrow range to visually isolate fine-grained policy divergence.}}

\label{fig:learning_curves_convergence}
\end{figure}

Figure~\ref{fig:learning_curves_convergence} illustrates the trials' averaged episodic returns for engagement, diversity, and fairness across the 100-episode training horizon. These learning curves capture the agents' behaviors under decaying $\epsilon$-greedy exploration, explicitly revealing the differences in policy convergence between single-objective scalarization and multi-objective Pareto optimization.

During training, the Standard DQN exhibits a characteristic pattern of progressive fairness degradation: the smoothed fairness return strictly decays from an initial plateau of roughly $118$ to $80$ at convergence. This degradation is a direct consequence of exposure concentration. Because the engagement-only agent repeatedly exploits a narrow manifold of high-affinity items, its cumulative exposure count $C(a_t)$ grows. This actively minimizes the fairness reward for those highly-exploited items while starving the long tail of any exploratory gradient signal.

By contrast, the Pareto-DQN jointly optimizes all three objectives via Hypervolume maximization, fundamentally mitigating this concentration effect. It initiates and maintains a substantially higher absolute fairness baseline throughout training, converging near $120$. 

Furthermore, the Pareto-DQN demonstrates a capacity to preserve semantic variance, the diversity dimension, middle pane of Figure~\ref{fig:learning_curves_convergence}. By strategically recommending semantically distant items, the agent continuously perturbs the user's preference centroid, effectively counteracting the homogenization characteristic of algorithmic filter bubbles. 


\subsection{Multi-Objective Trade-off Analysis}

\begin{figure}[htpb]
\centering
\adjustbox{trim=1.5cm 1cm 0.5cm 0.84cm, clip}{
\includesvg[width=0.58\textwidth]{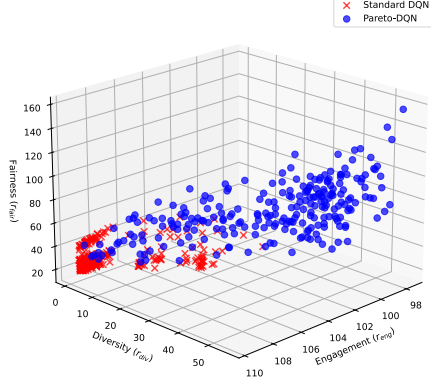}
}
\caption{Three-dimensional evaluation returns $(r_{eng}, r_{div}, r_{fair})$ for Standard DQN (red crosses) and Pareto-DQN (blue circles). Points represent the strictly non-dominated solutions extracted from 61 held-out test users aggregated across five independent trials. The Pareto-DQN maps a significantly broader region of the Pareto manifold, achieving dominant fairness and diversity metrics while sustaining comparable engagement.}
\label{fig:pareto3d}
\end{figure}

Figure~\ref{fig:pareto3d} presents a three-dimensional scatter plot of the evaluation returns $(r_{eng}, r_{div}, r_{fair})$ under the fully greedy policy ($\epsilon = 0$). To evaluate multi-objective performance, we aggregate the per-user cumulative returns across five independent trials. We apply a non-dominance sorting to the pooled evaluation trajectories to extract the final Pareto set. Each data point in Figure~\ref{fig:pareto3d} thus represents a verified, strictly non-dominated evaluation episode.

The Standard DQN cluster (red crosses) occupies a constrained, degenerate region of the objective space. Consistent with its single-objective design, it successfully drives engagement into the $100$-$110$ range but suffers from semantic collapse, yielding exceptionally low fairness ($<40$) and minimal diversity. Conversely, the Pareto-DQN cluster (blue circles) robustly populates a much broader Pareto manifold. It achieves higher fairness and more widely distributed diversity metrics, demonstrating the agent's capacity to balance structurally antagonistic objectives without catastrophic engagement degradation.

Crucially, the engagement axis exhibits considerable overlap between the two agents. For a clearer, simplified two-dimensional projection of this multi-objective trade-off surface, see Figure \ref{fig:price}. This sustained engagement is structurally governed by our stratified candidate-pooling mechanics: because $70\%$ of the candidate pool ($\rho = 0.3$) consists of KNN-retrieved items with high semantic affinity to the user's state centroid $\bm{s_t}$, the Pareto-DQN promotes fairness and diversity.

\subsection{Filter Bubble Mitigation via Semantic Variance}

To empirically quantify the filter bubble phenomenon, we analyze the variance in user embeddings. For every user evaluation (episode), it tracks their state over time $T$. Then we defined the trace of the state trajectory's covariance matrix, $\text{tr}(\text{Cov}([\bm{s_0}, \bm{s_1}, \ldots, \bm{s_T}]))$. A low trace indicates semantic homogenization (the operational signature of a filter bubble \cite{Nguyen2014FilterBubble}), whereas a high trace reflects a diverse sequence of recommendations that actively traverse a broader region of the embedding space. Our trajectory variance metric captures the long-term semantic mobility of the user profile across consecutive interaction cycles

\begin{figure}[htpb]
\centering
\adjustbox{trim=0cm 0cm 0cm 0.3cm, clip}{
\includesvg[width=0.39\textwidth]{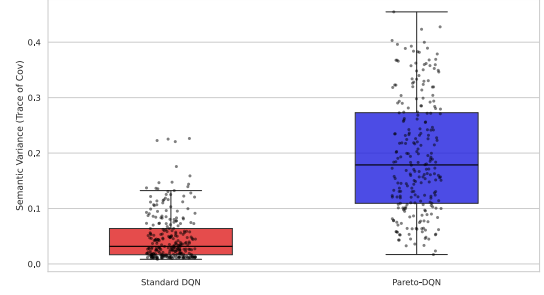}
}
\caption{Distribution of user embedding variance (trace of covariance matrix) across the extracted non-dominated sets. The Standard DQN (red) exhibits a highly compressed, illustrating severe semantic homogenization. Conversely, the Pareto-DQN (blue) sustains a broad, high-variance distribution (spanning up to $\sim\!0.45$), empirically demonstrating the disruption of the filter bubble.}
\label{fig:bubble}
\end{figure}

Figure~\ref{fig:bubble} reveals a structural divergence in the distribution of state trajectory variance between the non-dominated sets of the two agents. The Standard DQN exhibits a compressed variance distribution, with a median approaching $0.03$. By greedily maximizing engagement ($r_{eng}$), the single-objective baseline continuously exploits items with high semantic affinity to the current state $s_t$. 

In contrast, the Pareto-DQN agent successfully preserves state mobility, yielding a broad distribution with significantly higher variance, with a median near $0.18$ and observations extending up to $\sim\!0.45$. By explicitly optimizing the full vectorial reward $\mathbf{r}_t$, the hypervolume-based action selection actively negotiates the structural antagonism between engagement and diversity. This optimization effectively forces the periodic injection of semantically distant items into the user's exposure stream across its Pareto-optimal solutions. Consequently, the user's preference centroid is continually perturbed, empirically validating that our multi-objective Pareto formulation successfully disrupts the positive feedback loops that drive semantic collapse.

\subsection{Price of Responsibility}

\begin{figure}[htpb]
\centering
\adjustbox{trim=0cm 0cm 0cm 0.55cm, clip}{
\includesvg[width=0.43\textwidth]{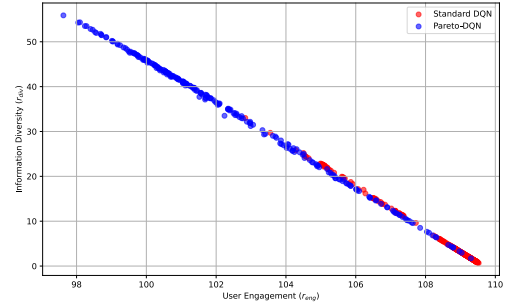}
}
\caption{Per-user evaluation returns projected onto the engagement and diversity plane. The linear decay confirms the antagonism between the objectives. The Standard DQN (red) collapses to the extreme engagement maximizing region, while the Pareto-DQN (blue) maps the full Pareto front, unlocking high diversity with a proportional and manageable engagement trade-off.}
\label{fig:price}
\end{figure}

The price of responsibility quantifies the theoretical engagement cost incurred when a system explicitly optimizes for auxiliary societal objectives like diversity and fairness \cite{Ge2022Pareto}. Figure~\ref{fig:price} projects the per-user evaluation returns onto the two aspects: engagement and diversity, exposing the strict geometry of this trade-off surface.

The scatter plot reveals a strict linear boundary, empirically verifying the structural antagonism between $r_{eng}$ and $r_{div}$. The Standard DQN (red) predictably collapses into the extreme lower-right region, maximizing engagement (up to 110) while severely suppressing diversity (predominantly below 30). Because the single-objective agent receives no gradient signal to explore semantically distant items, it remains structurally blind to the broader objective space. 

In contrast, the Pareto-DQN (blue) comprehensively maps the Pareto frontier. It successfully navigates into high-diversity regions (reaching up to 55) by making controlled, proportional sacrifices in engagement (scaling down smoothly to 98). This distribution demonstrates a highly manageable price of responsibility: the agent can achieve massive relative gains in diversity and fairness for only fractional absolute drops in engagement. 


\section{Conclusion and future works}\label{sec:conclusion}

This study establishes that formulating sequential recommendations as a multi-objective Markov decision process effectively mitigates the filter bubble phenomenon inherent to single-objective scalarization. By integrating high-fidelity semantic embeddings with a Pareto-DQN, our framework natively navigates the structural trade-offs among user engagement, information diversity, and provider fairness. Empirical evaluations in our offline simulation framework suggest that the Pareto-DQN effectively populates a broad region of the Pareto frontier, thereby disrupting the positive feedback loops that lead to semantic collapse under the simulated conditions. Unlike the single-objective baseline, our agent maintains high state-trajectory variance, unlocking massive relative gains in auxiliary societal objectives at only fractional, manageable drops in engagement. 


Beyond architectural evolution, we acknowledge the limitations of the current experimental setup. While the MovieLens-Small dataset serves as a robust benchmark for filter bubbles, future evaluations should encompass diverse domains such as e-commerce and news recommendation to further validate the framework's generalizability. Furthermore, future work will address the inherent assumptions of offline RL simulations by incorporating stochastic preference drift models and comparing the Pareto-DQN against additional diversity-aware baselines.

\section{Acknowledgments}
The authorship team would like to acknowledge the vision, support and guidance of the IEEE Industrial Electronics Society in conducting the Generative AI Hackathon under the leadership of Daswin De Silva and Lakshitha Gunasekara.

\bibliographystyle{IEEEtran}
\bibliography{references}

\end{document}